\documentclass[journal]{IEEEtran}

\usepackage[utf8]{inputenc}
\usepackage{siunitx}
\usepackage[colorlinks=true, allcolors=blue]{hyperref}
\usepackage{amsmath}
\usepackage{amsfonts}
\usepackage{soul} 
\usepackage{subcaption}
\usepackage{stfloats}
\usepackage{booktabs}
\usepackage{enumitem}

\usepackage{tikz,xcolor}

\usepackage[ruled,linesnumbered]{algorithm2e}
\usepackage{algpseudocode}

\definecolor{lime}{HTML}{A6CE39}
\DeclareRobustCommand{\orcidicon}
{
    \begin{tikzpicture}
    \draw[lime, fill=lime] (0,0) circle [radius=0.16] 
    node[white] {{\fontfamily{qag}\selectfont \tiny ID}};    \draw[white, fill=white] (-0.0625,0.095) circle [radius=0.007];    
    \end{tikzpicture}
    \hspace{0mm}}
\foreach \x in {A, ..., Z}{%
    \expandafter\xdef\csname orcid\x\endcsname{\noexpand\href{https://orcid.org/\csname orcidauthor\x\endcsname}{\noexpand\orcidicon}}
}

\setlist[enumerate]{itemsep = 0pt, parsep = 0pt, topsep = 0pt} 
\setlist[itemize]{itemsep = 0pt, parsep = 0pt, topsep = 0pt} 

\begin{document}

\title{Acceleration method for generating perception failure scenarios based on editing Markov process}

\author{Canjie Cai \vspace*{-30pt}
\thanks{Canjie Cai is with the Department of Information Engineering at Guangdong University of Technology, Guangzhou, China.}
}

\markboth{Journal of \LaTeX\ Class Files,~Vol.
}%
{Shell \MakeLowercase{\textit{et al.}}: Bare Demo of IEEEtran.cls for IEEE Journals}

\maketitle

\begin{abstract}
With the rapid advancement of autonomous driving technology, self-driving cars have become a central focus in the development of future transportation systems. 
Scenario generation technology has emerged as a crucial tool for testing and verifying the safety performance of autonomous driving systems. 
Current research in scenario generation primarily focuses on open roads such as highways, with relatively limited studies on underground parking garages. 
The unique structural constraints, insufficient lighting, and high-density obstacles in underground parking garages impose greater demands on the perception systems, which are critical to autonomous driving technology.

This study proposes an accelerated generation method for perception failure scenarios tailored to the underground parking garage environment, aimed at testing and improving the safety performance of autonomous vehicle (AV) perception algorithms in such settings. 
The method presented in this paper generates an intelligent testing environment with a high density of perception failure scenarios by learning the interactions between background vehicles (BVs) and autonomous vehicles (AVs) within perception failure scenarios. 
Furthermore, this method edits the Markov process within the perception failure scenario data to increase the density of critical information in the training data, thereby optimizing the learning and generation of perception failure scenarios.
A simulation environment for an underground parking garage was developed using the Carla and Vissim platforms, with Bevfusion employed as the perception algorithm for testing. 
The study demonstrates that this method can generate an intelligent testing environment with a high density of perception failure scenarios and enhance the safety performance of perception algorithms within this experimental setup.

\end{abstract}


%
\IEEEpeerreviewmaketitle

\section{Introduction}

As autonomous driving technology rapidly advances, self-driving cars have become a focal point in future transportation systems. 
One of the biggest challenges in autonomous driving research today is verifying the safety performance of autonomous vehicles. 
Existing testing methods include virtual simulation and real-world road tests in natural driving environments \cite{lan2023virtual}. 
Natural driving scenarios are characterized by high parameter dimensions, large parameter ranges, and massive combinations, with dangerous scenarios only constituting a small fraction, exhibiting a "long tail effect."
According to the findings in \cite{kalra2016driving}, demonstrating with 95\% confidence that autonomous vehicles can achieve a 20\% reduction in accident rates compared to human-driven vehicles necessitates at least 14.2 billion kilometers of testing in naturalistic road environments. 
Due to this severe inefficiency, autonomous vehicle developers face significant economic and time costs, which severely hinder the market entry process for autonomous vehicles.

Currently, scenario generation technology has become key to enhancing and ensuring the safety performance of autonomous driving systems. 
This technology primarily involves efficiently generating and simulating various driving scenarios to verify and improve the safety performance of autonomous driving systems, such as simulating traffic accidents, sudden weather conditions, and extreme traffic behaviors on open roads. 
Scenario generation technology also ensures that autonomous driving algorithms are tested under safe conditions, reducing the cost of testing these algorithms.

Scenario generation studies mainly focus on open roads such as highways, with relatively limited studies on underground parking garages. 
Underground parking garages, with their structural constraints, insufficient lighting, and high-density obstacles, pose higher demands on the critical perception systems of autonomous driving technology. 
The perception systems of autonomous vehicles face greater challenges in environments like underground parking garages, where any failure could lead to severe safety consequences.

This study proposes an accelerated generation method for perception failure scenarios tailored to underground parking garage environments, aimed at testing and improving the safety performance of perception algorithms in autonomous vehicles under such conditions. 
The method presented in this paper generates an intelligent testing environment with a high density of perception failure scenarios by learning the interactions between background vehicles (BVs) and autonomous vehicles (AVs) within perception failure scenarios. 
Furthermore, this method edits the Markov process within the perception failure scenario data to increase the density of critical information in the training data, thereby optimizing the learning and generation of perception failure scenarios.

The main contributions of this method can be summarized as follows:
\begin{enumerate}
\item Unlike previous methods for generating perception failure scenarios \cite{jain2019analyzing,tu2020physically}, this method generates perception failure scenarios by learning from the interactions between background vehicles (BVs) and autonomous vehicles (AVs).
\item It incorporates the editing Markov process from \cite{feng2023dense}, demonstrating its effectiveness in generating perception failure scenarios.
\item It provides a scenario generation scheme that enhances the safety performance of perception algorithms in underground parking garage environments.
\end{enumerate}

\section{Related Work}
\label{sec:related_work}

Scenario generation technology has become a key tool for validating and improving the safety performance of autonomous driving systems. 
This technology primarily involves the efficient generation and simulation of various driving scenarios to validate and improve the safety performance of autonomous systems, such as simulating traffic accidents, sudden weather conditions, and extreme traffic behaviors on open roads. 
According to \cite{ding2023survey,riedmaier2020survey}, current scenario generation methods can be broadly categorized into three types: data-driven, adversarial generation-based, and knowledge-based scenario generation methods.

Data-driven scenario generation methods \cite{van2015automated,kruber2018unsupervised,suo2021trafficsim,ehrhardt2020relate} mainly rely on replaying or learning from real-world driving data to generate scenarios. 
These methods produce scenarios with higher realism but lack controllability and efficiency when generating dangerous or other specific scenarios.
Adversarial generation-based scenario generation methods \cite{jain2019analyzing,tu2020physically,wachi2019failure} primarily produce scenarios by targeting the perception, decision-making, and control algorithms of autonomous systems, creating scenarios that these algorithms struggle to handle. 
These methods are more efficient in generating dangerous or specific scenarios, but the realism of the scenarios is lower.
Knowledge-based scenario generation methods \cite{lan2022semantic,bagschik2018ontology,yi2024key,liu2024mining} typically use expert knowledge and predefined rules to generate scenarios that meet specific standards. 
These methods offer higher controllability but lack realism in the generated scenarios.
Each method has unique characteristics and can be flexibly chosen or combined based on specific application requirements.

Current perception failure scenario generation methods \cite{jain2019analyzing,tu2020physically} are primarily based on adversarial generation techniques, attacking perception algorithms through the generation of adversarial images or point clouds to create perception failure scenarios. 
\cite{jain2019analyzing} proposed a method using differentiable rendering to generate images that induce errors in object detection \cite{lan2019evolving,lan2018real} or misclassification by classification algorithms \cite{lan2022class,gao2021neat}. 
\cite{tu2020physically} developed a method to generate universal 3D adversarial objects that deceive LIDAR detectors, causing point cloud-based object detection algorithms to malfunction.
The limitations of these methods can be summarized as follows:
\begin{enumerate}
\item The generated scenarios are primarily static, based on point clouds or images.
\item They mainly rely on adversarial generation techniques, making it difficult to ensure the realism of the scenarios.
\item There is a lack of generalizability, making it difficult to adapt to different types of perception algorithms.
\end{enumerate}

\section{Methodology}
\label{sec:methodology}

\subsection{Problem Description}
This section primarily introduces the problem description for accelerated generation of perception failure scenarios based on interactions between background vehicles (BVs) and autonomous vehicles (AVs). 
In this study, the entire scenario is denoted as $\boldsymbol{x}$, where $\boldsymbol{x}$ consists of the initial states of the AV and BVs and all the maneuvers across the time steps.
Specifically, $\boldsymbol{S}_k$ represents the state of all vehicles at time step $k$ (including position, velocity, etc.), and $\boldsymbol{U}_k$ denotes the maneuvers of all BVs at the same time step.
For individual vehicles, $\boldsymbol{s}_{kj}$ and $\boldsymbol{u}_{kj}$ respectively indicate the state and maneuver of the $j$-th BV at time step $k$. 
Notably, $\boldsymbol{s}_{k0}$ represents the state of the AV at time step $k$. 
The entire scenario can be expressed as $\boldsymbol{x} = [\boldsymbol{S}_0, \boldsymbol{U}_0, \boldsymbol{U}_1, \ldots, \boldsymbol{U}_T]$, where each $\boldsymbol{S}_k$ and $\boldsymbol{U}_k$ are the sets of vehicle states and maneuvers, respectively.

At each time step $k$, assuming $P_q$ represents the maneuver distribution for a single BV and that the maneuvers of different BVs are independent, the conditional probability of the set of all BV maneuvers $\boldsymbol{U}_k$, given the states of all BVs and the AV at time step $\boldsymbol{S}_k$, can be expressed as:
\begin{equation} \centering \small 
P(\boldsymbol{U}_k \mid \boldsymbol{S}_k)=\prod_{j=1}^{M} P_{q}\left(\boldsymbol{u}_{kj} \mid \boldsymbol{S}_k\right)
\end{equation}

Utilizing the Markov property, the probability distribution of the entire scenario can be represented as a product over the time steps:
\begin{equation} \centering \small 
P(\boldsymbol{x})=P(\boldsymbol{S}_0) \times \prod_{k=0}^{T} P(\boldsymbol{U}_k \mid \boldsymbol{S}_k)
\end{equation}

The probability of a perception failure event $A$ can be estimated using the Monte Carlo method, assuming a total number of sampled scenarios $N$, with the following approximation:
\begin{equation} \centering \small 
P(A)=E_{\boldsymbol{x} \sim P(\boldsymbol{x})}[P(A \mid \boldsymbol{x})] \approx \sum_{i=1}^{N} P\left(A \mid \boldsymbol{x}_{i}\right) P\left(\boldsymbol{x}_{i}\right)
\end{equation}

Expanding $P\left(\boldsymbol{x}_{i}\right)$ yields:
\begin{equation} \centering \small 
P(A) \approx \sum_{i=1}^{N} P\left(A \mid \boldsymbol{x}_{i}\right) P\left(\boldsymbol{S}_{i0}\right) \prod_{k=0}^{T} \prod_{j=1}^{M} P_{q}\left(\boldsymbol{u}_{ikj} \mid \boldsymbol{S}_{ik}\right)
\end{equation}

Thus, optimizing the maneuver behavior distribution $P_q$ of BVs to maximize $P(A)$ can increase the likelihood of occurrence of perception failure scenarios:
\begin{equation} \centering \small \label{eq:eq1}
\max _{P_{q}} P(A) \approx \max _{P_{q}} \sum_{i=1}^{N} P\left(A \mid \boldsymbol{x}_{i}\right) \prod_{k=0}^{T} \prod_{j=1}^{M} P_{q}\left(\boldsymbol{u}_{ikj} \mid \boldsymbol{S}_{ik}\right)
\end{equation}

\subsection{Editing Markov Process}
In autonomous driving data, non-safety-critical states predominate, with safety-critical states constituting only a small fraction. 
This distribution makes it challenging for traditional deep learning models to effectively learn from data cluttered with a majority of non-safety-critical states \cite{lan2022vision}. 
To address this issue, \cite{feng2023dense} introduced the concept of editing Markov process. 
By removing non-safety-critical states from the Markov process, training the model exclusively with data from safety-critical states significantly enhances the model’s ability to learn from complex scenarios.

For complex environments like underground parking garages, this paper also applies the method of editing Markov process to optimize the scenario generation process. 
Non-critical states, defined as states that minimally contribute to the final generation of hazardous scenarios, are excluded from the state set $\boldsymbol{x}_{C}$. Thus, it can be assumed that:
\begin{equation} \centering \small 
P\left(A \mid \boldsymbol{x}_{C}\right) \approx P(A \mid \boldsymbol{x})
\end{equation}

Let $\boldsymbol{S}_{C}$ represent the critical states, and use $I_{\boldsymbol{S}_{ik} \in \boldsymbol{S}_{C}}$ as the indicator function for critical states. 
The original optimization problem can be approximated as:
\begin{equation} \centering \small \label{eq:eq2}
\max _{P_{q}} P(A) \approx \max _{P_{q}} \sum_{i=1}^{N} P\left(A \mid \boldsymbol{x}_{Ci}\right) \prod_{k=0}^{T} \prod_{j=1}^{M} P_{q}\left(\boldsymbol{u}_{ikj} \mid \boldsymbol{S}_{ik}\right) I_{\boldsymbol{S}_{ik} \in \boldsymbol{S}_{C}}
\end{equation}

\subsection{Transforming the Deep Learning Problem}
This study employs deep learning methods to optimize the maneuver distribution $P_q$ of background vehicles (BVs) to maximize the probability of perception failure scenarios, $P(A)$. 
Based on \autoref{eq:eq2}, this paper sets the loss function as shown in \autoref{eq:eq3}:
\begin{equation} \centering \small \label{eq:eq3}
L = - \sum_{i=1}^{N} P\left(A \mid \boldsymbol{x}_{Ci}\right) \prod_{k=0}^{T} \prod_{j=1}^{M}
P_{q}\left(\boldsymbol{u}_{ikj} \mid \boldsymbol{S}_{ik}\right) I_{\boldsymbol{S}_{ik} \in \boldsymbol{S}_{C}}
\end{equation}

To further simplify, this paper uses the indicator function for perception failure events, $I_A(\boldsymbol{x})$, as an approximation for $P\left(A \mid \boldsymbol{x}_{Ci}\right)$, and transforms \autoref{eq:eq3} into:
\begin{equation} \centering \small \label{eq:eq4}
L = - \sum_{i=1}^{N} I_A(\boldsymbol{x}) \prod_{k=0}^{T} \prod_{j=1}^{M}
 P_{q}\left(\boldsymbol{u}_{ikj} \mid \boldsymbol{S}_{ik}\right) I_{\boldsymbol{S}_{ik} \in \boldsymbol{S}_{C}}
\end{equation}

Since the product operation might lead to issues with numerical stability during training, this paper ultimately approximates \autoref{eq:eq4} as:
\begin{equation} \centering \small 
L = - \sum_{i=1}^{N} I_A(\boldsymbol{x}) \sum_{k=0}^{T} \sum_{j=1}^{M}
 \log P_{q}\left(\boldsymbol{u}_{ikj} \mid \boldsymbol{S}_{ik}\right) I_{\boldsymbol{S}_{ik} \in \boldsymbol{S}_{C}}
\end{equation}

\subsection{Generating an Intelligent Testing Environment}
This section explains how the trained maneuver distribution model of background vehicles (BVs) is used to generate an intelligent testing environment with an increased number of perception failure scenarios. 
The method for generating the intelligent testing environment primarily modifies the calculation of the BVs' maneuver distributions based on the original environment. 
The specific process for computing the BVs' maneuvers in the intelligent testing environment is detailed in the pseudocode shown in \autoref{alg:bvs_maneuvers}.

\begin{algorithm}
\caption{Computation of BVs' maneuvers in an intelligent testing environment}
\label{alg:bvs_maneuvers}
\SetKwInOut{Input}{input}\SetKwInOut{Output}{output}
\SetKwData{AvState}{AvState}
\SetKwData{BvsStateArray}{BvsStateArray}
\SetKwData{BvsActionArray}{BvsActionArray}
\SetKwData{BvState}{BvState}
\SetKwData{BvAction}{BvAction}
\SetKwFunction{IsCriticalState}{IsCriticalState}
\SetKwFunction{ComputeActionFromModel}{ComputeActionFromModel}
\SetKwFunction{ComputeActionStandard}{ComputeActionStandard}
\Input{{State of the AV \AvState, All BV states \BvsStateArray}}
\Output{{All BV maneuvers \BvsActionArray}}

\BlankLine
\ForEach{\BvState in \BvsStateArray}{
    \eIf{\IsCriticalState{\BvState, \AvState}}{
        \tcp{If BV is in a critical state, compute the maneuver using the trained distribution model}
        
        \BvAction $\leftarrow$ \ComputeActionFromModel{\BvState, \AvState}\;
    }{
        \tcp{If BV is not in a critical state, compute the maneuver using the standard distribution}
        
        \BvAction $\leftarrow$ \ComputeActionStandard{\BvState, \AvState}\;
    }
    Append \BvAction to \BvsActionArray\;
}
\end{algorithm}

\section{Experiments}
\label{sec:experiments}

\subsection{Simulation Environment Setup}

\begin{figure}[!ht] \centering
    \includegraphics[width=0.98\linewidth]{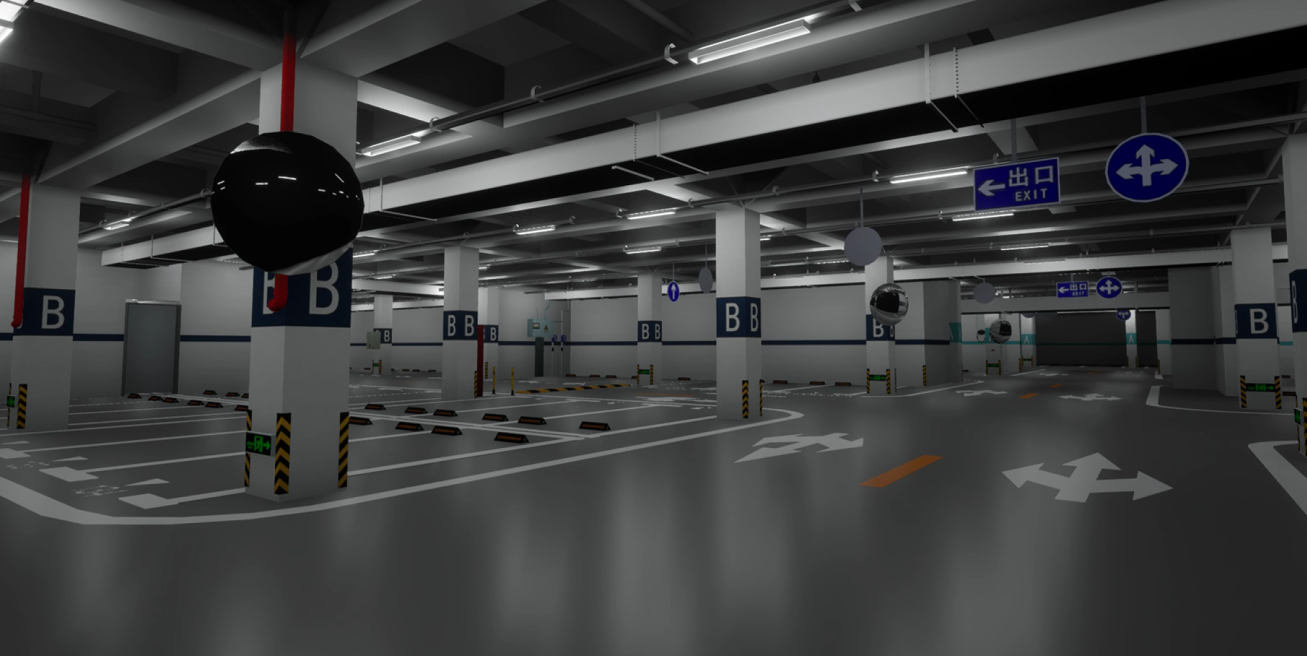}
    \caption{Carla simulation environment of the underground parking garage. The model is a 1:1 simulation of the underground parking garage at the Southern University of Science and Technology of Engineering, including lane markings, parking spaces, and traffic signs with semantic information.}\label{fig:parklot_static}
\end{figure}

\textbf{Static Simulation Environment}.
This study constructs a static simulation environment of an underground parking garage using Carla \cite{dosovitskiy2017carla}. 
The model utilized is sourced from the OpenAVP project and is based on a 1:1 simulation of the actual structure and environmental features of the underground parking garage at the Southern University of Science and Technology of Engineering. 
It perfectly replicates the details of the real environment, as illustrated in \autoref{fig:parklot_static}.

\begin{figure}[!ht] \centering
    \includegraphics[width=0.8\linewidth]{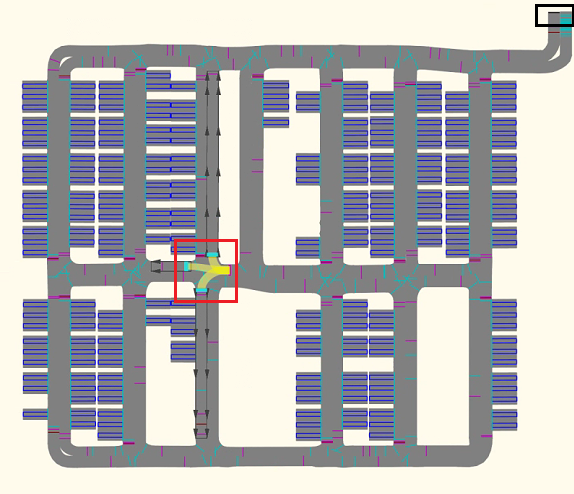}
    \caption{Vissim underground parking garage network. Red boxes are examples of path decision points in Vissim, where vehicles are assigned one of the three yellow paths shown upon passing these points. Black boxes indicate the vehicle entry and exit points; dark blue boxes represent parking spaces; gray areas denote the roads.}\label{fig:parklot_network}
\end{figure}

\textbf{Traffic Flow Generation}.
This research employs PTV Vissim \cite{fellendorf2010vissim} to develop the road network within the static simulation environment of the underground parking garage, subsequently generating the vehicle traffic. 
Specifically, decision points are added at all junctions within the network, allowing the study to influence vehicle routing decisions at these junctions by adjusting the decision points \cite{lan2023end}. 
The final configuration of the Vissim underground parking garage network is shown in \autoref{fig:parklot_network}.

\begin{figure}[!ht] \centering
    \includegraphics[width=0.98\linewidth]{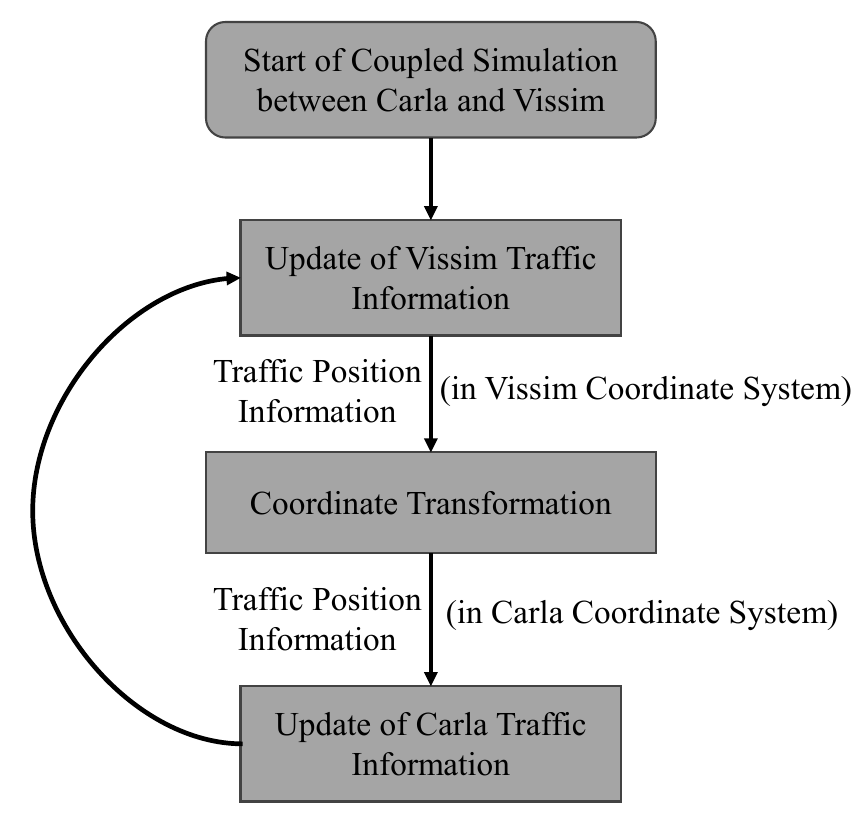}
    \caption{The coupled simulation process with Carla and Vissim.}
    \label{fig:carla_vissim_coupled_sim}
\end{figure}

\textbf{Coupled Simulation}.
This research generates the final underground parking garage simulation environment through a coupled simulation using Carla and PTV Vissim. 
The specific process of the Carla and Vissim coupled simulation is depicted in \autoref{fig:carla_vissim_coupled_sim}, including the following technical aspects:
\begin{enumerate}
\item Using Vissim's COM interface to read vehicle position data.
\item Constructing a Bridge module to translate vehicle position data from the Vissim coordinate system to the Carla coordinate system.
\item Synchronizing Vissim traffic within Carla.
\end{enumerate}

\begin{figure}[!ht] \centering
    \includegraphics[width=0.9\linewidth,trim={10 20 10 10},clip]{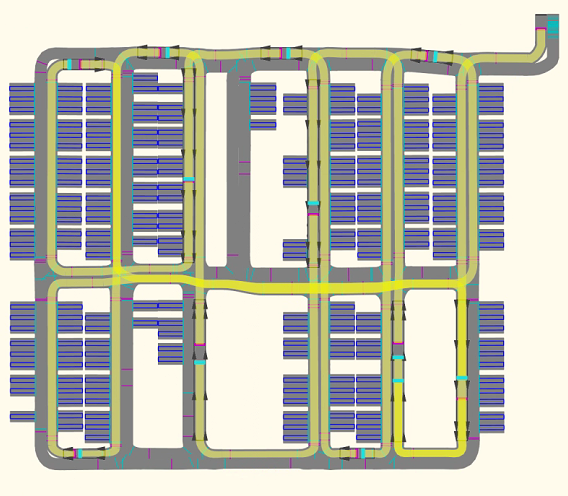}
    \caption{AV driving route (highlighted in yellow).}\label{fig:parklot_AV_route}
\end{figure}

\textbf{Simulation Environment Configuration}.
In this study, the autonomous vehicle (AV) navigates a fixed route within the underground parking garage simulation environment, as depicted in \autoref{fig:parklot_AV_route}. 
The sensor configurations of the AV used in this study are detailed in \autoref{tab:av_camera_config} and \autoref{tab:av_lidar_config}.

\begin{table}[!ht] \centering
    \begin{tabular}{c|c|c|c}
        \toprule
        Sensor Name & Position (x, y, z) & yaw & FOV \\
        \midrule
        CAM\_FRONT & (1.5, 0, 2) & 0°  & 70° \\
        \midrule
        CAM\_FRONT\_RIGHT & (1.5, 0.7, 2) & 55° & 70° \\
        \midrule
        CAM\_FRONT\_LEFT & (1.5, -0.7, 2) & -55° & 70° \\
        \midrule
        CAM\_BACK & (-1.5, 0, 2) & 180°  & 110° \\
        \midrule
        CAM\_BACK\_LEFT & (-0.7, -1.5, 2) & -110°  & 70° \\
        \midrule
        CAM\_BACK\_RIGHT & (-0.7, 1.5, 2) & 110° &  70° \\
        \bottomrule
    \end{tabular}
    \caption{AV surround camera configuration.}
    \label{tab:av_camera_config}
\end{table}

\begin{table}[!ht] \centering
    \begin{tabular}{c|c|c|c}
        \toprule
        Sensor Name & Position (x, y, z) & HFOV & VFOV Range \\
        \midrule
        LIDAR\_TOP & (1.5, 0, 2) &  360° & (-30°, 10°) \\ 
        \bottomrule
    \end{tabular}
    \caption{AV top lidar configuration. HFOV is horizontal field of view. VFOV Range is vertical field of view range.}
    \label{tab:av_lidar_config}
\end{table}

\subsection{Deployment of Perception Algorithm}

\textbf{Selection of Perception Algorithm}.
This study employs Bevfusion \cite{liu2023bevfusion} as the perception algorithm for subsequent experimental testing. 
Bevfusion is a multimodal 3D object detection algorithm that fuses data from cameras and LiDAR \cite{xu2019online}. 
The main reasons for choosing Bevfusion include:
\begin{enumerate}
\item Its ability to fuse camera and LiDAR data, which is particularly effective in handling structurally complex environments such as underground parking garages.
\item Its widespread validation in both academic and industrial contexts.
\item Support from an open-source framework.
\end{enumerate}

\textbf{Data Collection}.
For this study, a dataset of an underground parking garage was collected following the format used by nuscenes \cite{nuscenes2019}. 
Specific details about the dataset are provided in \autoref{tab:dataset_details}.

\begin{figure}[!ht] \centering
    \includegraphics[width=0.95\linewidth,trim={15 13 10 0},clip]{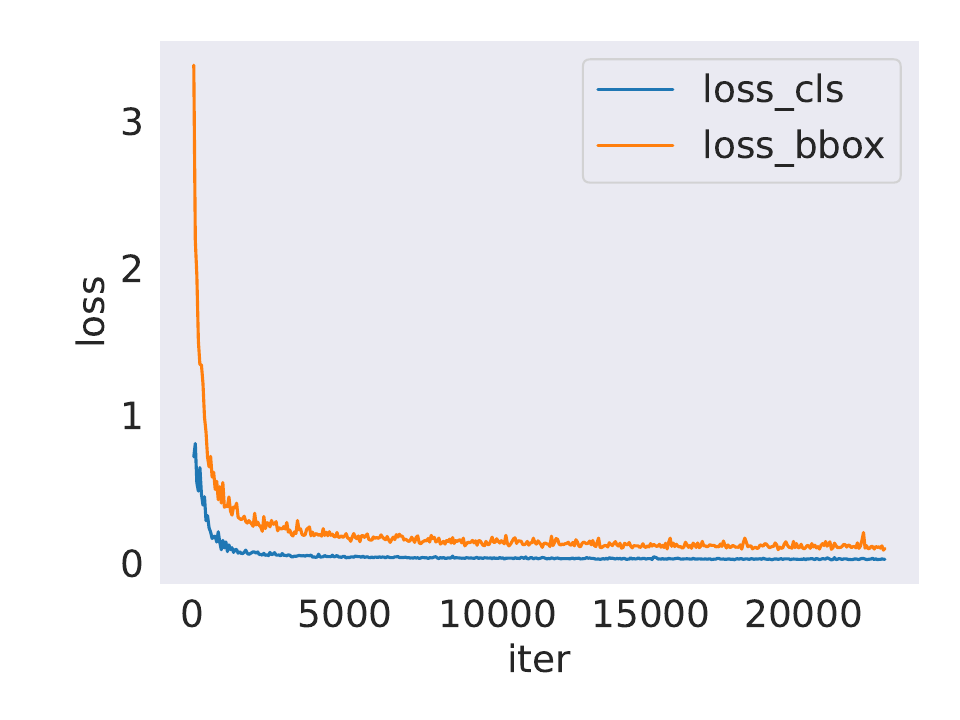}
    \caption{Loss curve during Bevfusion training.}\label{fig:bevfusion_loss}
\end{figure}

\begin{figure}[!ht] \centering
    \includegraphics[width=0.98\linewidth]{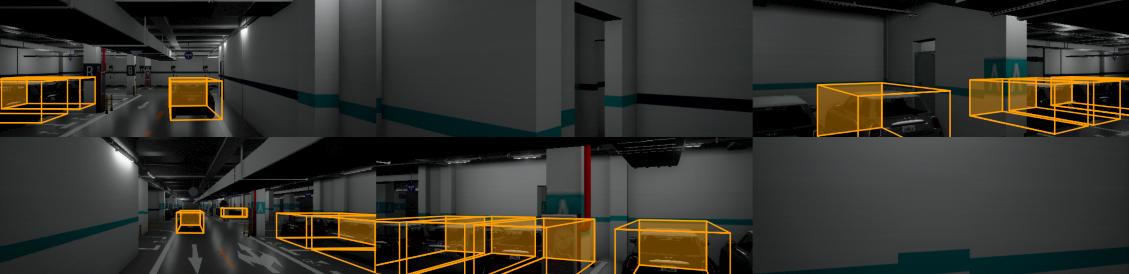}
    \includegraphics[width=0.98\linewidth]{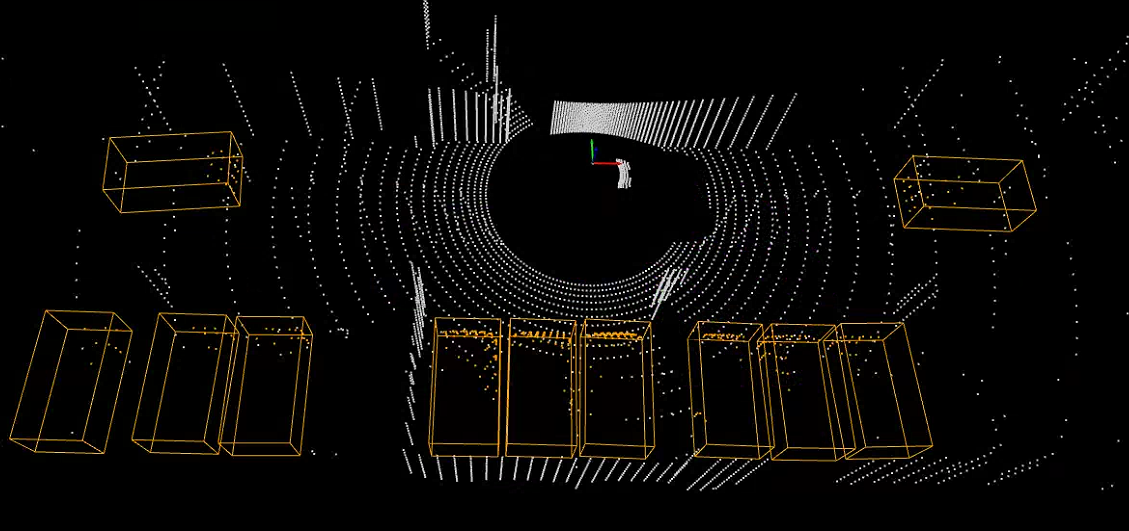}
    \caption{Visualization of Bevfusion's vehicle detection in an underground parking garage. The upper image shows pictures captured by the six surround cameras mounted on the AV; the lower image displays the point cloud from the AV's top-mounted LiDAR. The orange boxes indicate vehicles detected by the model.}\label{fig:bevfusion_infer}
\end{figure}

\textbf{Retraining of Bevfusion}.
The hardware configuration and training parameters used to train Bevfusion are detailed in \autoref{tab:bevfusion_hw_config} and \autoref{tab:bevfusion_train_params}, respectively. The loss curve during the Bevfusion training is shown in \autoref{fig:bevfusion_loss}. The inference results post-Bevfusion training are illustrated in \autoref{fig:bevfusion_infer}.

\begin{table}[!ht] \centering
    \begin{tabular}{c|c|c|c|c}
        \toprule
        Freq. & Dur. & Frames & Ann. & Format \\
        \midrule
        2Hz & 950s & 1900 & 3D vehicle boxes & nuscenes \cite{nuscenes2019} \\
        \bottomrule
    \end{tabular}
    \caption{Details of the underground parking garage dataset. Freq. is Sampling Frequency. Dur. is Collection Duration. Frames is Total Frames. Ann. is Annotations, including 3D vehicle bounding boxes. Format is Data Format.}
    \label{tab:dataset_details}
\end{table}

\begin{table}[!ht] \centering
    \begin{tabular}{c|c}   \toprule
        Hardware Type & Specifications \\  \midrule
        CPU & Intel\ Core\ i7-8700\ @\ 3.2GHz $\times$ 12 \\ \midrule
        GPU & NVIDIA\ GeForce\ GTX\ 1080\ 8GB \\  \midrule
        RAM & 32GB\ DDR4 \\  \bottomrule
    \end{tabular}
    \caption{Hardware configuration for Bevfusion training.}
    \label{tab:bevfusion_hw_config}
\end{table}

\begin{table}[!ht] \centering
    \begin{tabular}{c|c|c|c|c}  \toprule
        Optimizer & LR & Weight Decay & Batch Size & Epochs \\   \midrule
        AdamW & 0.00005 & 0.01 & 1 & 12 \\  \toprule
    \end{tabular}
    \caption{Bevfusion training parameters.}
    \label{tab:bevfusion_train_params}
\end{table}

\subsection{Experimental Setup}

\textbf{Validation of the Effectiveness of Accelerated Generation of Perception Failure Scenarios}.
This experiment aims to validate the proposed method's effectiveness for accelerating the generation of perception failure scenarios. 
The main steps and settings of this experiment are as follows:
\begin{enumerate}
\item Collect driving scenario data from the simulation environment. 
This experiment collected 17 hours of driving scenario data at a frequency of 2Hz, totaling 122,400 frames, with each frame's data details shown in \autoref{tab:frame_data_details}.
\item Select perception failure scenario data. 
This experiment uses the data frames from the 10 seconds prior to each perception failure frame as a perception failure scenario dataset.
\item Identify critical states within the perception failure scenario data. 
In this experiment, visible BVs within a 20-meter radius of the AV in the perception failure frame are considered critical BVs. 
All states of critical BVs within 7.5 seconds before the failure are set as critical states.
\item Train the BV maneuver distribution model using only critical states versus using all data. 
The input to the BV maneuver distribution model includes the states of the AV and BVs, and the output is the BV’s maneuvers, with specific input-output details referring to \autoref{tab:frame_data_details}. 
The Training parameters for the BV maneuver distribution model are shown in \autoref{tab:bv_model_train_params}.
\item Compare the proportion of perception failure frames within 0.5 hours in the generated intelligent testing environment.
\end{enumerate}

\textbf{Validation of Enhanced Safety Performance of the Perception Algorithm}.
The purpose of this experiment is to validate that the intelligent testing environment generated by the proposed method can enhance the safety performance of the perception algorithm. The main steps and settings of this experiment are:
\begin{enumerate}
\item Collect a dataset in the intelligent testing environment generated by the proposed method for retraining Bevfusion, keeping the dataset specifics consistent with  \autoref{tab:dataset_details}.
\item Retrain Bevfusion using the dataset collected in the intelligent testing environment. The hardware configuration and training parameters for training Bevfusion are consistent with \autoref{tab:bevfusion_hw_config} and \autoref{tab:bevfusion_train_params}.
\item Compare the proportion of perception failure frames within 0.5 hours between the original simulation environment and the intelligent testing environment using the trained Bevfusion.
\end{enumerate}

It is noteworthy that to verify the generalizability of the proposed method, this study conducted repeated experiments using different definitions of perception failures. 
The relationship between experiment numbers and definitions of perception failures is shown in \autoref{tab:failure_defs_by_exp}.

\begin{table}[!ht] \centering
    \begin{tabular}{c|c} \toprule
        Attribute & Description \\   \midrule
        AV Status & Position information of the AV \\ \midrule
        BVs Status & Position information of the BVs \\ \midrule
        BVs Maneuvers & Path choices at decision points by BVs \\ \midrule
        Perception Data & Detected vehicle positions by Bevfusion \\ \bottomrule
    \end{tabular}
    \caption{Specific contents of single frame data in scenario.}
    \label{tab:frame_data_details}
\end{table}

\begin{table}[!ht] \centering
    \begin{tabular}{c|c|c|c} \toprule
        Optimizer & LR & Batch Size & Epochs \\ \midrule
        Adam & 0.001 & 200000 & 400 \\ \bottomrule
    \end{tabular}
    \caption{Training parameters for the BV maneuver distribution model.}
    \label{tab:bv_model_train_params}
\end{table}

\begin{table}[!ht] \centering
    \begin{tabular}{c|c} \toprule
        Exp. No. & Definition of Perception Failure \\ \midrule
        a & $TE_{max} > 0.5$  \\ \midrule
        b & $TE_{max} > 0.8$  \\ \midrule
        c & $TE_{max} > 1.0$   \\ \midrule
        d & $FN > 0$  \\ \bottomrule
    \end{tabular}
    \caption{Definitions of perception failure corresponding to experiment numbers. $TE_{max}$ represents the maximum positional deviation between Bevfusion's predicted bounding boxes and the ground truth, and $FN$ represents the number of missed detections of actual vehicles by Bevfusion.}
    \label{tab:failure_defs_by_exp}
\end{table}

\section{Results}
\label{sec:results}

\begin{figure}[!ht] \centering
\begin{subfigure}[b]{0.47\linewidth}
    \includegraphics[width=\linewidth,trim={30 20 10 0},clip]{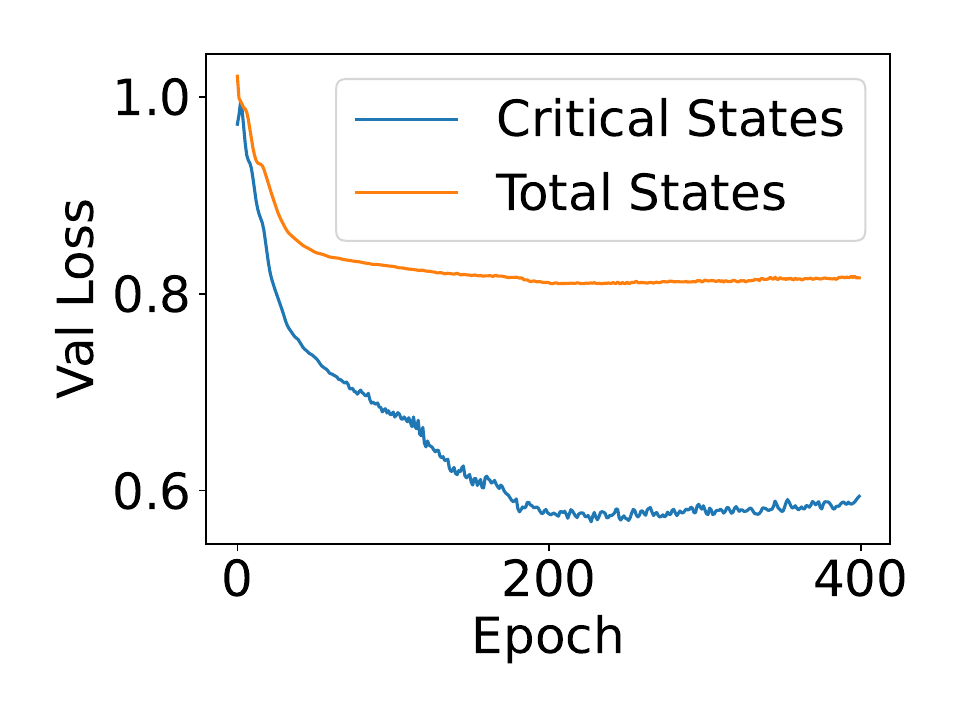}
    \caption{}
\end{subfigure}
\begin{subfigure}[b]{0.47\linewidth}
    \includegraphics[width=\linewidth,trim={30 20 10 0},clip]{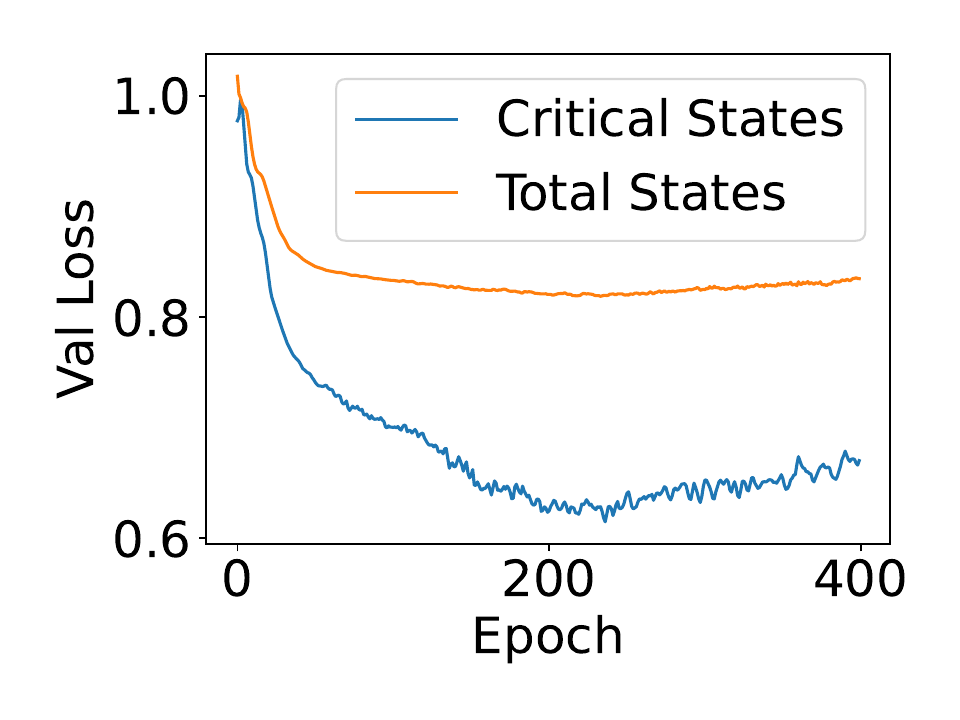}
    \caption{}
\end{subfigure}

\begin{subfigure}[b]{0.47\linewidth}
    \includegraphics[width=\linewidth,trim={30 20 10 0},clip]{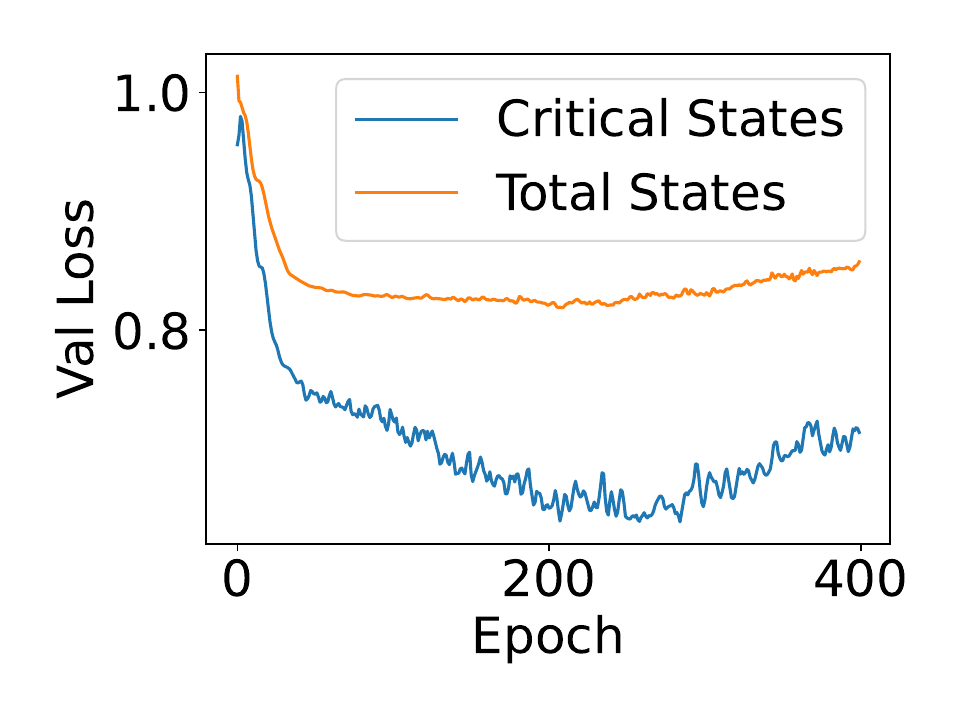}
    \caption{}
\end{subfigure}
\begin{subfigure}[b]{0.47\linewidth}
    \includegraphics[width=\linewidth,trim={30 20 10 0},clip]{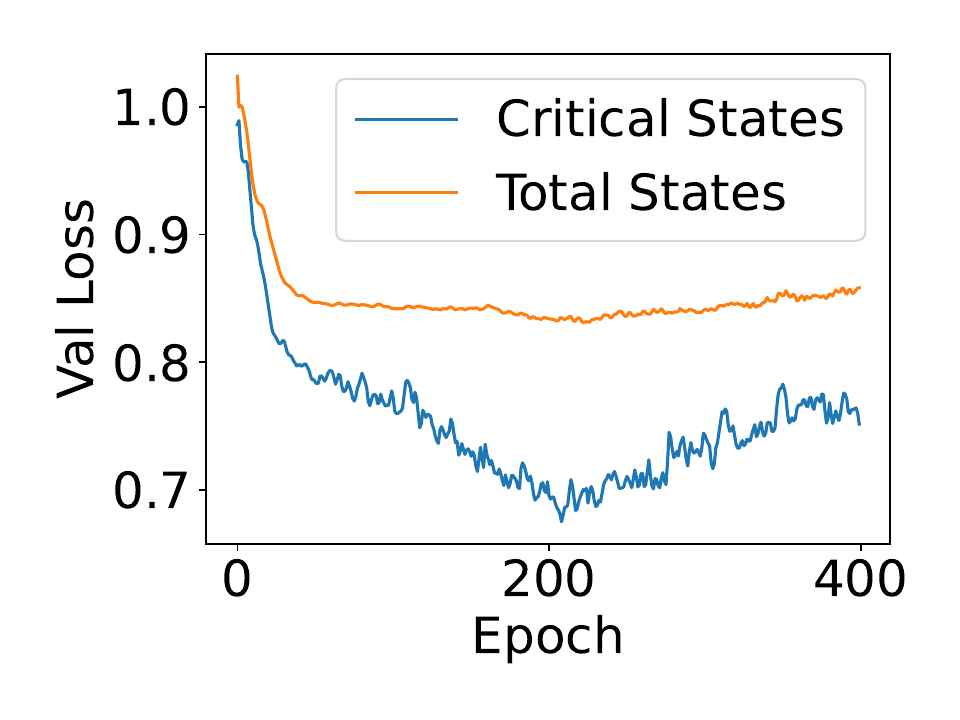}
    \caption{}
\end{subfigure}
\caption{Comparison of validation loss curves for training the BV maneuver distribution model using only critical states versus using all states.}\label{fig:val_loss_exp}

\end{figure}

\textbf{Validation of the Effectiveness of Accelerated Generation of Perception Failure Scenarios}.
Based on \autoref{tab:states_num_comparison}, it can be observed that the number of critical states in perception failure scenario data is significantly lower than the total number of states. 
From \autoref{fig:val_loss_exp}, it is evident that training the BV maneuver distribution model using only critical states from perception failure scenarios results in a faster decrease in validation set loss. 
As shown in \autoref{tab:failure_frame_ratios_comparison_env}, the proportion of perception failure frames in intelligent testing environment a is similar to that in the original simulation environment, whereas environment b shows a much higher proportion of failure frames.
These results validate that the proposed method can generate an intelligent testing environment enriched with high-density perception failure scenarios and confirm the effectiveness of the editing Markov process in optimizing the learning \cite{lan2022time,lan2021learning,lan2021learning2} and generation of these scenarios. 
Notably, \autoref{fig:val_loss_exp} reveals noticeable overfitting in all experiments except for experiment a, which has a higher volume of data as seen in \autoref{tab:states_num_comparison}. 
Therefore, the overfitting observed in \autoref{fig:val_loss_exp} is hypothesized to be due to insufficient data volume.

\begin{figure}[!ht] \centering
    \includegraphics[width=0.9\linewidth, height=0.4\linewidth]{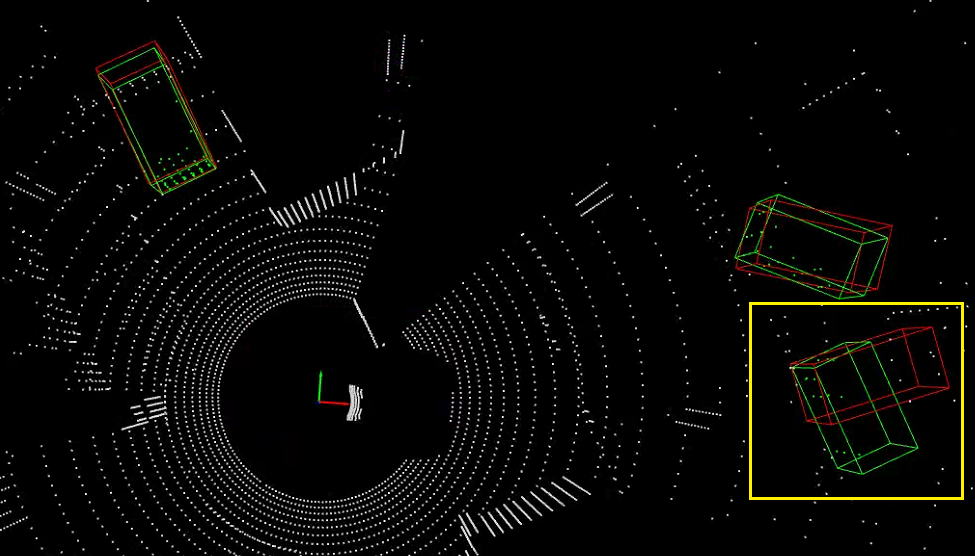}
    \includegraphics[width=0.9\linewidth, height=0.4\linewidth]{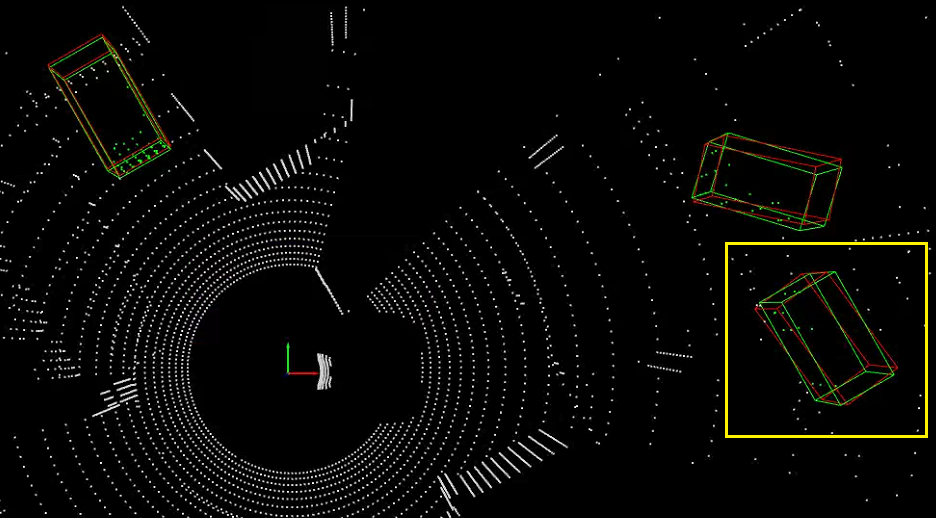}
    \caption{Comparison of Bevfusion vehicle detection performance trained on different datasets in Experiment c. The upper and lower images show the vehicle detection performance of Bevfusion trained on the original simulation environment dataset and the intelligent testing environment dataset, respectively. Green and red boxes represent the ground truth and model prediction boxes, respectively. Yellow boxes highlight areas with significant differences in recognition performance between the two models.}\label{fig:bevfusion_perf_exp_c}
\end{figure}

\begin{figure}[!ht] \centering
    \includegraphics[width=0.9\linewidth]{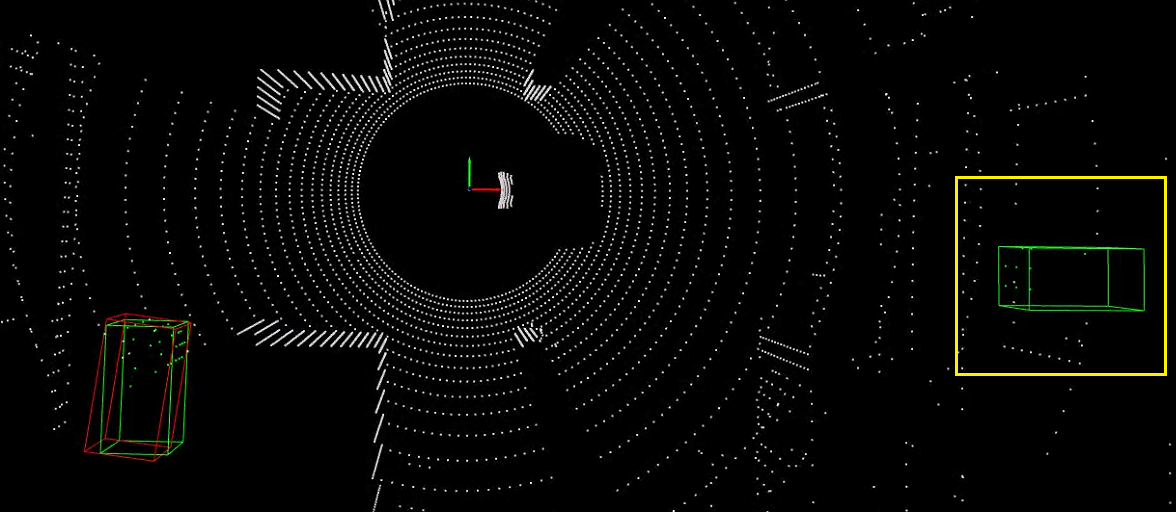}
    \includegraphics[width=0.9\linewidth]{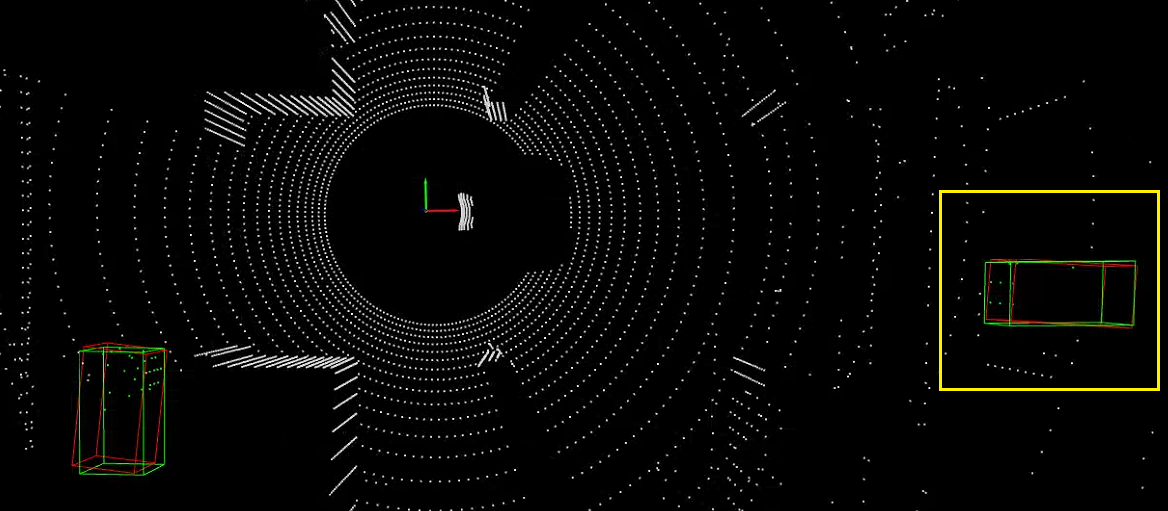}
    \caption{Comparison of Bevfusion vehicle detection performance trained on different datasets in Experiment d. }\label{fig:bevfusion_perf_exp_d}
\end{figure}

\textbf{Validation of Enhanced Safety Performance of the Perception Algorithm}.
According to \autoref{tab:failure_frame_ratios_comparison_dataset}, it is observed that datasets collected in the intelligent testing environment significantly reduce the proportion of perception failure frames of Bevfusion in the original simulation environment compared to those collected under normal conditions \cite{lan2019simulated}. 
Some visual comparison results are shown in \autoref{fig:bevfusion_perf_exp_c} and \autoref{fig:bevfusion_perf_exp_d}. 
It is speculated that the increased proportion of perception failure frames in the intelligent testing environment enables Bevfusion to handle these scenarios more effectively. 
Overall, these results validate that the proposed method can enhance the safety performance of the perception algorithm in the original simulation environment.

\begin{table}[!ht] \centering
    \begin{tabular}{ccc} \toprule
        Exp. No. & Total States & Critical States \\ \midrule
        a   &  106024 & 17926    \\
        b   &  46618  & 7941     \\
        c   &  22553  & 3783     \\
        d   &  26563  & 4669     \\ \bottomrule
    \end{tabular}
    \caption{Comparison of total states and critical states in perception failure scenario data.}
    \label{tab:states_num_comparison}
\end{table}

\begin{table}[!ht] \centering
    \begin{tabular}{cccc}     \toprule
        Exp. No. & Orig. Sim. Env. & Int. Test Env. a & Int. Test Env. b \\  \midrule
        a  & 5.53\%  & 6.42\% & 15.37\% \\
        b  & 2.14\%  & 2.25\% & 4.95\%  \\
        c  & 1.11\%  & 1.56\% & 3.31\%  \\
        d  & 1.33\%  & 1.25\% & 2.72\%  \\    \bottomrule
    \end{tabular}
    \caption{Comparison of perception failure frame ratios across different environments. Exp. No. is Experiment Number; Orig. Sim. Env. is Original Simulation Environment. Int. Test Env. a is Intelligent Testing Environment a, generated using the BV maneuver distribution model trained with all states from the perception failure scenario data. Int. Test Env. b is Intelligent Testing Environment b, generated using the BV maneuver distribution model trained only with critical states from the perception failure scenario data.}
    \label{tab:failure_frame_ratios_comparison_env}
\end{table}

\begin{table}[!ht] \centering
    \begin{tabular}{ccc} \toprule
        Exp. No. & Orig. Sim. Env. & Int. Test Env.  \\ \midrule
        a  & 5.53\% & 2.64\%                  \\
        b  & 2.14\% & 1.00\%                  \\
        c  & 1.11\% & 0.50\%                  \\
        d  & 1.33\% & 0.89\%                  \\ \bottomrule
    \end{tabular}
    \caption{Comparison of the proportion of perception failure frames in the original simulation environment when Bevfusion is trained on different datasets. The Orig. Sim. Env. refers to the dataset collected in original simulation environment. The Int. Test Env. refers to the dataset collected in intelligent testing environment.}
    \label{tab:failure_frame_ratios_comparison_dataset}
\end{table}

\section{Discussion}
\label{sec:discussion}

The results of this study validate the proposed method's ability to accelerate the generation of perception failure scenarios and enhance the safety performance of perception algorithms. 
Compared to the perception failure scenario generation methods mentioned in \autoref{sec:related_work}, the proposed method exhibits the following characteristics:
\begin{enumerate}
\item It is data-driven and can extend from simulated to real-world scene data, enhancing the realism of the generated scenarios.
\item The method is not designed for a specific perception algorithm, allowing for better scalability across different perception technologies.
\item It generates scenarios through interactions between BVs and AVs, rather than directly producing point clouds and images, which facilitates further research into the impact of vehicle interactions on the generation of perception failure scenarios.
\end{enumerate}

\subsection{Limitations}

The method proposed in this paper has the following limitations:
\begin{enumerate}
\item Poor controllability of scenario generation. It is unable to generate specific types of perception failure scenarios in a controlled manner.
\item The method for selecting critical states in perception failure scenarios significantly impacts the performance of the proposed method.
\item It requires a large dataset of perception failure scenarios.
\end{enumerate}

\section{Conclusions}
\label{sec:conclusion}

The results of this study confirm that the proposed method can generate an intelligent testing environment with high-density perception failure scenarios based on the original simulation environment, thereby enhancing the safety performance of perception algorithms within that environment. 
Furthermore, the results also validate the effectiveness of the editing Markov process in optimizing the learning and generation of perception failure scenarios.

\bibliographystyle{IEEEtran}
\bibliography{bibliography}

\end{document}